\documentclass[oribibl,final]{llncs}

\usepackage{graphicx}
\usepackage{color}
\usepackage{tabularx}
\usepackage{achicago}
\usepackage{abbrevs}
\usepackage[T1]{url}

\let\em\itswitch

\newif\ifpdf
\ifx\pdfoutput\undefined \else
  \ifx\pdfoutput\relax
  \else
    \ifcase\pdfoutput
    \else
      \pdftrue
    \fi
  \fi
\fi

\providecommand{\ie}{\textit{i.e.} }%
\providecommand{\eg}{\textit{e.g.} }%
\providecommand{\possessivecite}[1]{\citeANP{#1}'s~\citeyear{#1}}%

\newabbrev\tdt{Topic Detection and Tracking (TDT)}[TDT]
\newabbrev\mds{Multi-document Summarization}[MDS]
\newabbrev\rst{\emph{Rhetorical Structure Theory}}[RST]
\newabbrev\cst{Cross-do\-cu\-ment Structure Theory (CST)}[CST]
\newabbrev\nlg{Natural Language Generation (NLG)}[NLG]
\newabbrev\IE{Information Extraction (IE)}[IE]
\newabbrev\nerc{Named Entity Recognition and Classification (NERC)}[NERC]
\newabbrev\nes{Named Entities (NEs)}[NEs]
\newabbrev\muc{Message Understanding Conferences (MUC)}[MUC]
\newabbrev\ml{Machine Learning}[ML]

\begin{document}

\title{An Introduction to the Summarization of Evolving Events: Linear and
Non-linear Evolution}
\author{Stergos D. Afantenos\inst{1} \and Konstantina Liontou\inst{2} \and
        Maria Salapata\inst{2} \and Vangelis Karkaletsis\inst{1}}
\institute{Software and Knowledge Engineering Laboratory\\
           Institute of Informatics and Telecommunications,\\
           National Center for Scientific Research (NCSR) ``Demokritos''
           \email{\{stergos,vangelis\}@iit.demokritos.gr}
           \and
           Institute of Language and Speech Processing}

\maketitle

\begin{abstract}
This paper examines the summarization of events that evolve through time. It
discusses different types of evolution taking into account the time in which
the incidents of an event are happening and the different sources reporting on
the specific event. It proposes an approach for multi-document summarization
which employs ``messages'' for representing the incidents of an event and
cross-document relations that hold between messages according to certain
conditions. The paper also outlines the current version of the summarization
system we are implementing to realize this approach.
\end{abstract}

\ResetAbbrevs{All}

\section{Introduction}
The exchange of information is of outmost importance for humans. Through the
history of humankind it has taken many forms, from gossiping to the publication
of news through dedicated media. More recently, the Internet has given a new
perspective to this human faculty, making the exchange of information much more
easy and virtually unrestricted.

Naturally this has caused some problems. Imagine, for example, that someone
wants to keep track of an event that is being described on various news
sources, over the Internet, as it evolves through time. The problem is that
there exist a plethora of news sources making very difficult for someone to
compare the different versions of the story in each source. Automatic text
summarization is a solution to this information overflow problem. In this paper
we propose a general framework for the automatic summarization of evolving
events, \ie the summarization of events that evolve through time.

A crucial question, that can possibly arise at this point, concerns the
definition of the ``event''. In the \tdt research an event is described as
``something that happens at some specific time and place''
(\citeNP{Papka99:PhD}, p 3; see also \citeNP{Allan&al98:TDTFinRep}). The
inherent notion of time is what distinguishes the event from the more general
term \emph{topic}. For example, incidents which include hostages are regarded
as topics, while a particular incident, such as the one concerning the two
Italian women that were kept as hostages by an Iraqi group in 2004, is regarded
as an event. In our discussion about ``events'' we will adopt this definition
provided by the \tdt research.

In the \mds community, a consensus that has emerged is that in order to
summarize a set of related documents, one has to identify similarities and
differences among the documents (\citeNP{Mani&Bloedorn99,Mani01}; see also
\citeNP{Endres-Niggemeyer98} and \citeNP{Afantenos&al.04:MedicalSurvey}). Yet,
no consensus has been reached concerning as to where those similarities and
differences should be targeted. In our work we propose that the similarities
and differences, at least for evolving events, should be viewed under two
perspectives: \emph{time} and \emph{source}, through \emph{cross-document
relations}. We call \emph{synchronic relations} those relations that are
concerned with the similarities and differences, between the various sources,
on the same temporal horizon and \emph{diachronic relations} those relations
that are more concerned with the evolution of an event as it is being described
by one source.

Summarization of evolving events should not be confused with evolving
summaries. Evolving summaries were originally proposed, but not implemented, by
\citeN[p 149]{Radev99:PhD} as follows: ``An evolving summary $S_{k+1}$ is the
summary of a story, numbered $A_{k+1}$, when the stories numbered $A_1$ to
$A_k$ have already been processed and presented in a summarized form to the
user. Summary $S_{k+1}$ differs from its predecessor, $S_k$, because it
contains new information and omits information from $S_k$''. What we propose,
instead, is a framework which will enable the creation of summaries of evolving
events.

Section~\ref{sec:kinds} discusses the different kinds of evolution in terms of
the time the incidents of an event are happening and in terms of the rate with
which the various news sources are emitting their reports.
Section~\ref{sec:msgs} introduces the notion of messages which we use for
representing the various incidents of an event. Section~\ref{sec:rels}
discusses the two types of cross-document relations (synchronic and diachronic)
which hold between messages. Section~\ref{sec:comp} outlines the system
developed so far that realizes our approach, as well as other options we are
currently investigating.

\section{Kinds of Evolution}\label{sec:kinds}
This work studies the summarization of events that evolve through time, as they
are being described by various sources. In this study we came to the conclusion
that we should distinguish between the evolution of an event in \emph{time} and
the \emph{rate} of reporting about an evolving event from various sources.

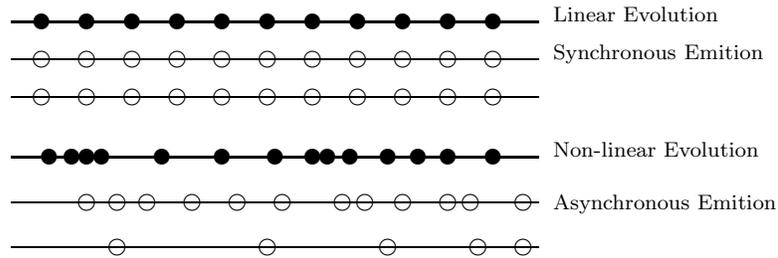
\begin{figure}[thb]
  \begin{center}
    \setlength{\unitlength}{4144sp}%
\begingroup\makeatletter\ifx\SetFigFont\undefined%
\gdef\SetFigFont#1#2#3#4#5{%
  \reset@font\fontsize{#1}{#2pt}%
  \fontfamily{#3}\fontseries{#4}\fontshape{#5}%
  \selectfont}%
\fi\endgroup%
\begin{picture}(4390,1493)(259,-1644)
{\color[rgb]{0,0,0}\thinlines
\put(496,-1051){\circle*{90}}
}%
{\color[rgb]{0,0,0}\put(631,-1051){\circle*{90}}
}%
{\color[rgb]{0,0,0}\put(721,-1051){\circle*{90}}
}%
{\color[rgb]{0,0,0}\put(811,-1051){\circle*{90}}
}%
{\color[rgb]{0,0,0}\put(1171,-1051){\circle*{90}}
}%
{\color[rgb]{0,0,0}\put(1846,-1051){\circle*{90}}
}%
{\color[rgb]{0,0,0}\put(2071,-1051){\circle*{90}}
}%
{\color[rgb]{0,0,0}\put(2161,-1051){\circle*{90}}
}%
{\color[rgb]{0,0,0}\put(2296,-1051){\circle*{90}}
}%
{\color[rgb]{0,0,0}\put(271,-1051){\line( 1, 0){2250}}
}%
{\color[rgb]{0,0,0}\put(2521,-1051){\line( 1, 0){900}}
}%
{\color[rgb]{0,0,0}\put(1531,-1051){\circle*{90}}
}%
{\color[rgb]{0,0,0}\put(2701,-1051){\circle*{90}}
}%
{\color[rgb]{0,0,0}\put(2881,-1051){\circle*{90}}
}%
{\color[rgb]{0,0,0}\put(3151,-1051){\circle*{90}}
}%
{\color[rgb]{0,0,0}\put(2521,-1051){\circle*{90}}
}%
{\color[rgb]{0,0,0}\put(451,-466){\circle{90}}
}%
{\color[rgb]{0,0,0}\put(721,-466){\circle{90}}
}%
{\color[rgb]{0,0,0}\put(991,-466){\circle{90}}
}%
{\color[rgb]{0,0,0}\put(1261,-466){\circle{90}}
}%
{\color[rgb]{0,0,0}\put(1531,-466){\circle{90}}
}%
{\color[rgb]{0,0,0}\put(1801,-466){\circle{90}}
}%
{\color[rgb]{0,0,0}\put(2071,-466){\circle{90}}
}%
{\color[rgb]{0,0,0}\put(2341,-466){\circle{90}}
}%
{\color[rgb]{0,0,0}\put(2611,-466){\circle{90}}
}%
{\color[rgb]{0,0,0}\put(2881,-466){\circle{90}}
}%
{\color[rgb]{0,0,0}\put(3151,-466){\circle{90}}
}%
{\color[rgb]{0,0,0}\put(3151,-691){\circle{90}}
}%
{\color[rgb]{0,0,0}\put(2881,-691){\circle{90}}
}%
{\color[rgb]{0,0,0}\put(2611,-691){\circle{90}}
}%
{\color[rgb]{0,0,0}\put(2341,-691){\circle{90}}
}%
{\color[rgb]{0,0,0}\put(2071,-691){\circle{90}}
}%
{\color[rgb]{0,0,0}\put(1801,-691){\circle{90}}
}%
{\color[rgb]{0,0,0}\put(1531,-691){\circle{90}}
}%
{\color[rgb]{0,0,0}\put(1261,-691){\circle{90}}
}%
{\color[rgb]{0,0,0}\put(991,-691){\circle{90}}
}%
{\color[rgb]{0,0,0}\put(721,-691){\circle{90}}
}%
{\color[rgb]{0,0,0}\put(451,-691){\circle{90}}
}%
{\color[rgb]{0,0,0}\put(451,-241){\circle*{90}}
}%
{\color[rgb]{0,0,0}\put(721,-241){\circle*{90}}
}%
{\color[rgb]{0,0,0}\put(991,-241){\circle*{90}}
}%
{\color[rgb]{0,0,0}\put(1261,-241){\circle*{90}}
}%
{\color[rgb]{0,0,0}\put(1531,-241){\circle*{90}}
}%
{\color[rgb]{0,0,0}\put(1801,-241){\circle*{90}}
}%
{\color[rgb]{0,0,0}\put(2071,-241){\circle*{90}}
}%
{\color[rgb]{0,0,0}\put(2341,-241){\circle*{90}}
}%
{\color[rgb]{0,0,0}\put(2611,-241){\circle*{90}}
}%
{\color[rgb]{0,0,0}\put(2881,-241){\circle*{90}}
}%
{\color[rgb]{0,0,0}\put(3151,-241){\circle*{90}}
}%
{\color[rgb]{0,0,0}\put(721,-1321){\circle{90}}
}%
{\color[rgb]{0,0,0}\put(901,-1321){\circle{90}}
}%
{\color[rgb]{0,0,0}\put(1081,-1321){\circle{90}}
}%
{\color[rgb]{0,0,0}\put(1351,-1321){\circle{90}}
}%
{\color[rgb]{0,0,0}\put(1891,-1321){\circle{90}}
}%
{\color[rgb]{0,0,0}\put(1621,-1321){\circle{90}}
}%
{\color[rgb]{0,0,0}\put(2251,-1321){\circle{90}}
}%
{\color[rgb]{0,0,0}\put(2386,-1321){\circle{90}}
}%
{\color[rgb]{0,0,0}\put(2611,-1321){\circle{90}}
}%
{\color[rgb]{0,0,0}\put(2881,-1321){\circle{90}}
}%
{\color[rgb]{0,0,0}\put(3016,-1321){\circle{90}}
}%
{\color[rgb]{0,0,0}\put(3331,-1321){\circle{90}}
}%
{\color[rgb]{0,0,0}\put(901,-1591){\circle{90}}
}%
{\color[rgb]{0,0,0}\put(1801,-1591){\circle{90}}
}%
{\color[rgb]{0,0,0}\put(2521,-1591){\circle{90}}
}%
{\color[rgb]{0,0,0}\put(3061,-1591){\circle{90}}
}%
{\color[rgb]{0,0,0}\put(3331,-1591){\circle{90}}
}%
{\color[rgb]{0,0,0}\put(2476,-466){\line( 1, 0){945}}
}%
{\color[rgb]{0,0,0}\put(2476,-691){\line( 1, 0){945}}
}%
{\color[rgb]{0,0,0}\put(2476,-241){\line( 1, 0){945}}
}%
{\color[rgb]{0,0,0}\put(2521,-1321){\line( 1, 0){900}}
}%
{\color[rgb]{0,0,0}\put(2521,-1591){\line( 1, 0){900}}
}%
{\color[rgb]{0,0,0}\put(271,-466){\line( 1, 0){2205}}
}%
{\color[rgb]{0,0,0}\put(271,-691){\line( 1, 0){2205}}
}%
{\color[rgb]{0,0,0}\put(271,-241){\line( 1, 0){2205}}
}%
{\color[rgb]{0,0,0}\put(271,-1321){\line( 1, 0){2250}}
}%
{\color[rgb]{0,0,0}\put(271,-1591){\line( 1, 0){2250}}
}%
\put(3511,-241){\makebox(0,0)[lb]{\smash{{\SetFigFont{8}{9.6}{\familydefault}{\mddefault}{\updefault}{\color[rgb]{0,0,0}Linear Evolution}%
}}}}
\put(3511,-1051){\makebox(0,0)[lb]{\smash{{\SetFigFont{8}{9.6}{\familydefault}{\mddefault}{\updefault}{\color[rgb]{0,0,0}Non-linear Evolution}%
}}}}
\put(3511,-466){\makebox(0,0)[lb]{\smash{{\SetFigFont{8}{9.6}{\rmdefault}{\mddefault}{\updefault}{\color[rgb]{0,0,0}Synchronous Emition}%
}}}}
\put(3511,-1366){\makebox(0,0)[lb]{\smash{{\SetFigFont{8}{9.6}{\rmdefault}{\mddefault}{\updefault}{\color[rgb]{0,0,0}Asynchronous Emition}%
}}}}
\end{picture}%
    \caption{Linear and Non-linear evolution}\label{fig:evolution}
  \end{center}
\end{figure}

Concerning the evolution of an event we distinguish between two types of
evolution: \emph{linear} and \emph{non-linear} evolution. In linear evolution
the major incidents of an event are happening in constant and possibly
predictable quanta of time. This means that if the first incident $q_0$ happens
at time $t_0$, then each subsequent incident $q_n$ will come at time $t_n = t_0
+ n*t$, where $t$ is the constant amount of time with which the incidents are
happening. In non-linear evolution, in contrast, we cannot distinguish any
meaningful pattern in the order that the major incidents of an event are
happening. This distinction is depicted in Figure~\ref{fig:evolution} in which
the evolution of two different events is depicted with the dark solid circles.

Linearly evolving events have a fair proportion in the world. They are related
with human activities which occur at regular intervals. One such example can be
the descriptions of various athletic events which occur regularly. In
particular we have examined the descriptions of football matches
\cite{Afantenos&al.04:SETN}. On the other hand, one can argue that most of the
events that we find in the news stories are non-linearly evolving events. They
can vary from political ones, such as elections or various international
political issues, to airplane crashes or terroristic events. Currently we are
investigating the domain of incidents which involve hostages.

In terms of the reporting on an event from various sources we can distinguish
between \emph{synchronous} and \emph{asynchronous} emission of reports. This
distinction is depicted in Figure~\ref{fig:evolution} with the white circles.
In most of the cases, when we have an event that evolves linearly we will also
have a synchronous emission of reports, since the various sources can easily
adjust to the pattern of the evolution of an event. This cannot be said for the
case of non-linear evolution, resulting thus in asynchronous emission of
reports by the various sources.

\begin{figure}[thb]
\begin{center}
  \ifpdf
    \includegraphics[scale=0.37]{images/linear.pdf}
    \includegraphics[scale=0.37]{images/nonlinear.pdf}
  \else
    \includegraphics[scale=0.37]{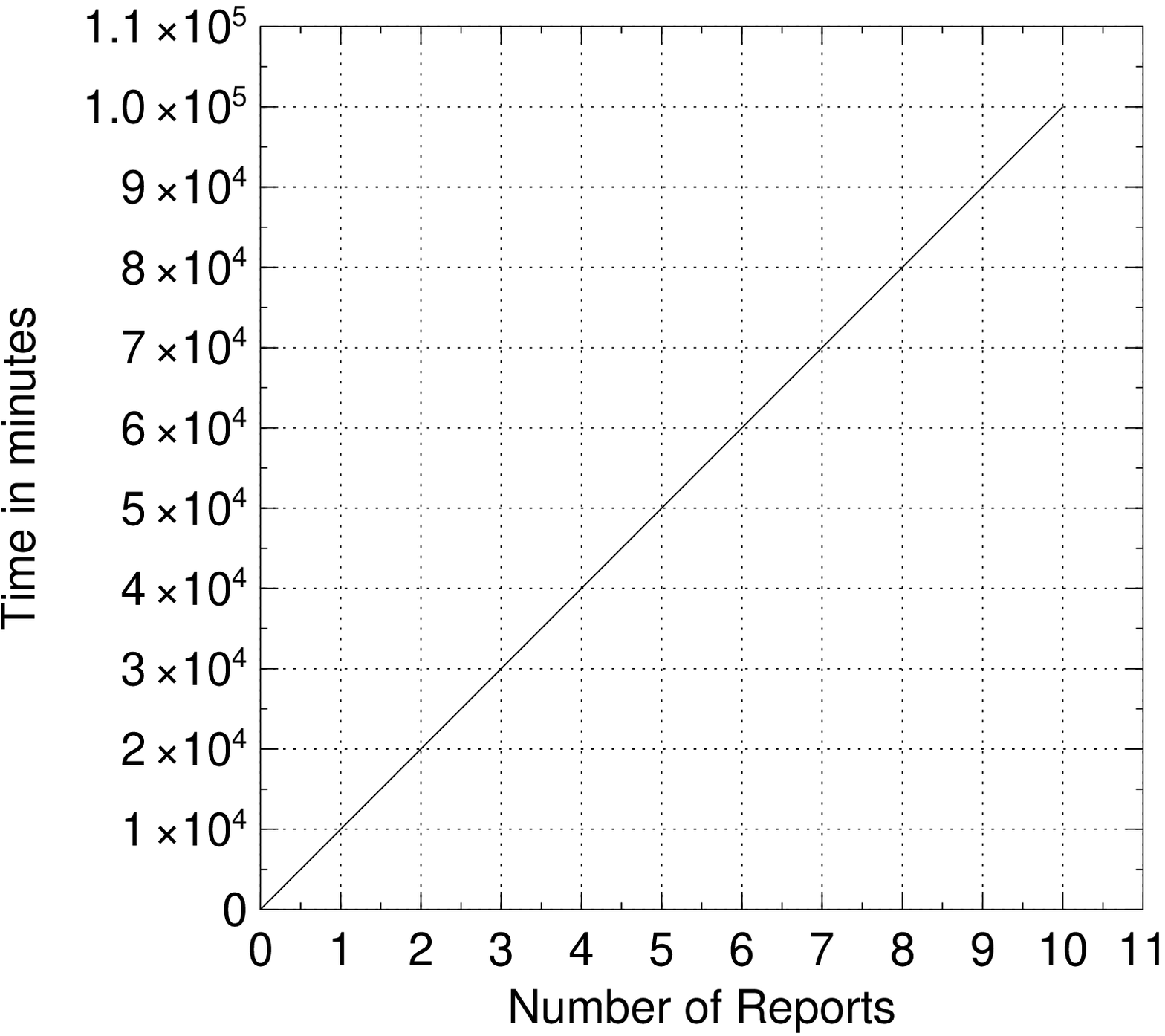}
    \includegraphics[scale=0.37]{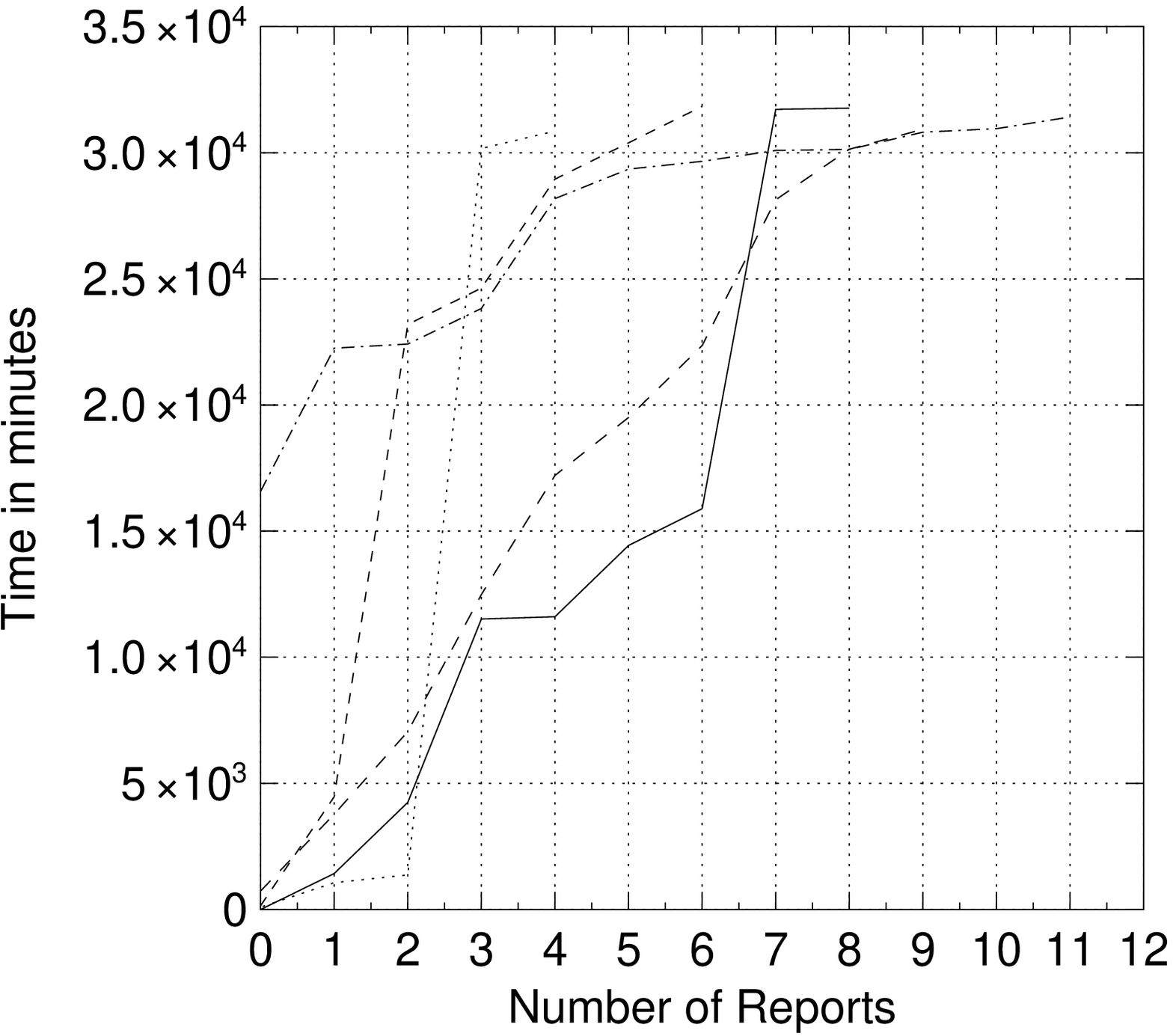}
  \fi
  \caption{Linear and Non-linear evolution}\label{fig:linearAndNonLinear}
\end{center}
\end{figure}

In Figure~\ref{fig:linearAndNonLinear} we represent two events which evolve
linearly and non-linearly and for which the sources report synchronously and
asynchronously respectively. The horizontal axis in this figure represents the
number of reports per source on a particular event. The vertical axis
represents the time, in minutes, that the documents are published. The first
event concerns descriptions of football matches. In this particular event we
have constant reports weekly, \ie every 10800 minutes, from 3 different
sources. The lines for each source fall on top of each other since they publish
simultaneously. The second event concerns a terroristic group in Iraq which
kept as hostages two Italian women threatening to kill them, unless their
demands were fulfilled. In the figure we depict 5 sources. The number of
reports that each source is making varies from five to twelve, in a period of
time of about 23 days. As we can see from the figure, most of the sources begin
reporting almost instantaneously, except one which delays its report for about
twelve days. Another source, although it reports almost immediately, it delays
considerably later reports.

The linearity or non-linearity of an evolving event, as well as the rate of
sources emission, affects our summarization approach which is based on the
exploitation of the similarities and differences that exist synchronically and
diachronically between the documents. The cross-document relations, and the way
that they are affected by linearity, will be explained in more detail in
section~\ref{sec:rels}. In the following section we will concentrate on the
notion of messages for representing the incidents of an event.

\section{Messages}\label{sec:msgs}
Each event is composed from various simpler incidents. For example, in the
football domain, such incidents can be the performance of a player or a team,
the goals that are achieved, the possible injuries of players, etc. In a domain
with hostages such incidents can be the occupation of a building, the
negotiations, the demands of the terrorists, the fact that they freed a
hostage, etc.

We use messages to represent those incidents. Each message is composed of two
parts: its \emph{type} and a list of \emph{arguments} which take their values
from an \emph{ontology} for the specific domain:\footnote{See
\citeN{Afantenos&al.04:SETN}.}
\begin{center}
  \texttt{message\_type ( arg$_1$, $\ldots$ , arg$_n$ )}\\
  where \texttt{arg}$_i$ $\in$ Domain Ontology
\end{center}
The message type represents the type of the incident, whilst the arguments
represent the main entities that are involved in this incident. It is possible
that some messages may be accompanied by some \emph{constraints} on their
arguments, which reflect various pragmatic constraints. These messages are
similar structures (although simpler ones) with the templates used in the
\muc.\footnote{\url{http://www.itl.nist.gov/iaui/894.02/related_projects/muc/proceedings/muc_7_toc.html}}

Each message is also linked to a specific source and time. In other words, if
we have a message $m$, then we have associated with it two extra pieces of
information, $m\texttt{.time}$ and $m\texttt{.source}$. Concerning the
\texttt{source}, it is inherited by the document that contains the message.
This cannot be said for the \texttt{time} as well, since the time of the
incidents might be different from the emission time. This is expressed in the
document by a temporal expression. Thus, in order to determine the
\texttt{time} of a message we should interpret this expression in relation to
the time of the publication of the document.

\begin{quote}\label{table:messages}
\begin{centering}
  \begin{tabular}{|l|l|}
    \hline
    \textbf{\textit{Linear}} & \textbf{\textit{Non-linear}}\\
    \hline\hline\scriptsize
    \textbf{performance} (entity, in\_what, time\_span, value) &
    \scriptsize\textbf{negotiate} (entity$_1$, entity$_2$, about)\\
    \begin{tabularx}{2.2in}{Xll}
      entity     &:& Player or Team\\
      in\_what   &:& Action Area\\
      time\_span &:& Minute or Duration\\
      value      &:& Degree
    \end{tabularx} &
    \begin{tabularx}{1.1in}{Xll}
      entity$_1$ &: Person\\
      entity$_2$ &: Person\\
      about      &: Activity
    \end{tabularx}\\ 
    \hline
  \end{tabular}
  \normalsize
\end{centering}
\end{quote}

Examples of messages' specifications, for a linear and a non-linear domain are
shown in the above table. The arguments for each message come from the domain
ontology. Thus, for example, the \texttt{Activity} argument in the second
message corresponds to a set of activities which are defined in the ontology of
the domain. The specifications for the first message come from the domain of
football matches \cite{Afantenos&al.04:SETN} and it represents the performance
of a player or a team for a specific time-span and a specific action area (\eg
in the defense). The specifications of the second message come from the topic
which is related with hostages, which we currently investigate. This message
represents the fact that we have a negotiation between two entities concerning
a specific activity (\eg the release of some hostages).

\section{Cross-document Relations}\label{sec:rels}
Cross-document relations hold between messages and are distinguished into
\emph{synchronic} and \emph{diachronic}.

Synchronic relations try to identify the similarities and differences that two
sources have, at about the same time. In the case of linear or synchronous
evolution all the sources report in the same time. Thus in most of the cases
the incidents described in each document refer to the time that the article was
published. Yet, in some cases we might have temporal expressions in the text
that modify the time that a message might refer. In such cases, before
establishing a synchronic relation, we should place this message in the
appropriate time horizon. In the case of non-linear asynchronous evolution this
phenomenon is predominant. Each source reports at irregular intervals, possibly
mentioning incidents that happened long before the publication of the article,
and which another source might have already mentioned in an article published
earlier (see the second part of Figure~\ref{fig:linearAndNonLinear}). In this
case we shouldn't rely any more to the publication of an article, but instead
on the \emph{time} tag that the messages have, which has been appropriately
modified according to the temporal expressions found in the text. Once this has
been performed, we should then establish a \emph{time window} in which we
should consider the messages, and thus the relations, as synchronic. This time
window, depending on the domain, can vary from some hours to some days.

Diachronic relations, on the other hand, try to capture the similarities and
differences, through time, that exist for an event as it is being described by
the \emph{same} source. In this sense, diachronic relations do not exhibit the
problems of time that the synchronic relations do.

Cross-document relations, in our viewpoint, are domain dependent, since they
represent pragmatic information which depends on the domain.\footnote{This does
not mean that we do not believe that domain independent relations could not
possibly exist. An example could be the relations agreement and disagreement,
which can obviously be independent of domain.} Examples of synchronic relations
can be agreement, disagreement, elaboration, generalization, etc. Examples of
diachronic relations can be positive or negative graduation, stability,
continuation, repetition, etc.

In more formal terms, if we represent a relation $r$ as a pair of messages
$\langle m_1, m_2 \rangle$, where $m_1$ and $m_2$ are two messages, then a
relation will be synchronic iff
\[m_1\texttt{.time} = m_2\texttt{.time} \textrm{ and }
  m_1\texttt{.source} \neq m_2\texttt{.source} \]
and diachronic iff
\[m_1\texttt{.time} > m_2\texttt{.time} \textrm{ and }
  m_1\texttt{.source} = m_2\texttt{.source} \]
We have to note that a relation has a directionality. As is evident,
diachronically a relation can hold from a past time to a future time. In the
case of a synchronic relation (\eg agreement) a relation can have both
directions, in which case we have in fact two relations.

In order to define a relation in a domain we have to provide a \emph{name} for
it, and describe the conditions under which it will hold. The name of the
relation is in fact \emph{pragmatic} information, which we will be able to
exploit during the generation of the summary. The conditions under which a
relation between two messages holds are represented in terms of values of their
arguments, as well as their corresponding time and source.

Suppose, for example, that we have two identical messages. If they have the
same temporal tag, but belong to different sources, then we have an
\textsl{agreement} relation. If, on the other hand, they have the same source
but chronological distance one or higher, then we can speak, for example, of a
\textsl{stability} relation. Thus we see that, apart from the characteristics
that the arguments of a message pair $\langle m_1, m_2 \rangle$ should exhibit,
the source and temporal distance also play a role for that pair to be
characterized as a relation.

\begin{figure}[htb]
  \centering
  \ifpdf
    \includegraphics{images/relations-example.pdf}
  \else
    \includegraphics{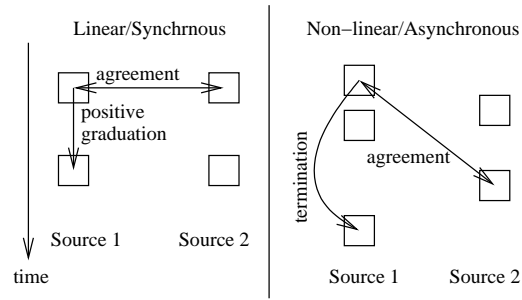}
  \fi
  \caption{Examples of synchronic and diachronic relations}
  \label{fig:rel-example}
\end{figure}

In Figure~\ref{fig:rel-example} we can see the difference, in terms of
synchronic relations, between a domain which evolves linearly and has a
synchronous emission of reports and a domain which evolves non-linearly and has
asynchronous emission of reports. In the first case we have two identical
\texttt{performance} messages (see the table of page~\pageref{table:messages}),
from two documents which have been published at the same time. Thus, and
according to the specifications of the synchronic relations
\cite{Afantenos&Karkaletsis.04}, we have an \textsl{agreement} relation. In the
second case we have two identical \texttt{negotiate} messages from documents
that have different publication times. Yet, in the text that defines those
messages, we have a temporal expression which modifies the \texttt{time} tag
for one of the messages, making them refer on the same day. Thus, again we have
an \textsl{agreement} relation, although the documents which contain the
messages have not been published on the same day.

In the same figure you can see two diachronic relations. In the linearly
evolving case we have two \texttt{performance} messages
\begin{center}
\small%
\texttt{performance (entity$_1$, in\_what$_1$, time\_span$_1$, value$_1$)}\\%
\texttt{performance (entity$_2$, in\_what$_2$, time\_span$_2$, value$_2$)}%
\normalsize%
\end{center}
which have identical arguments, except that \texttt{value$_1$ < value$_2$}. In
this case, and according to the specifications for the relations of the domain
\cite{Afantenos&al.04:SETN} we have a \textsl{positive graduation} diachronic
relation. In the second case we have two different messages
\begin{center}
\small%
\texttt{start (entity$_1$, activity$_1$)}\\%
\texttt{end (entity$_2$, activity$_2$)}%
\normalsize%
\end{center}
where \texttt{entity$_1$ = entity$_2$} and \texttt{activity$_1$ =
activity$_2$}. In this case, according to the specifications, we have a
\textsl{termination} diachronic relation. Note that in the first case we have a
diachronic relation that holds between the same message types, while in the
second case the diachronic relation holds between different message types.
Also, in the first case the documents that contain the messages have distance
one, \ie the one follows immediately the other, while in the second case they
have greater distance.

There may be also cases where an event is being described by one source but not
from the others. Since we need at least two messages from different sources in
order to have a synchronic relation, we will not connect that message with
another one, thus possibly missing an important piece of information that a
source is reporting. An \textsl{ellipsis} relation could be introduced to
handle such cases.

\section{Potential Computational Approaches}\label{sec:comp}
An initial study of a linearly evolving domain is presented in
\citeN{Afantenos&al.04:SETN}. In \citeN{Afantenos&Karkaletsis.04} we present a
system which automatically extracts the messages and the relations from the
text. The messages extraction sub-system involves two processing stages, one
for the identification of the messages' types and one for the filling in of its
arguments. During the first stage a classifier is trained. The word lemmas and
the Named Entities are used in the training vectors. The argument filling is
performed using heuristics. The sub-system implementing the extraction of
relations exploits the conditions under which a relation holds, as described in
the specifications of each relation.

Currently we are investigating a topic which evolves non-linearly with
asynchronous emission of reports, namely that of incidents involving hostages.
For this topic, apart from performing the above experiments concerning the
extraction of the messages and the relations, we are also implementing an
algorithm which identifies the various temporal expressions in the text. This
is essential since, as we have noted in
sections~\ref{sec:msgs}~and~\ref{sec:rels} in order to identify the synchronic
relation in a non-linearly evolving domain with asynchronous emission of
reports, we should not rely anymore on the time an article was published.
Instead we should recognize the time that a message is referring to, according
to the temporal expressions which characterize this message.

Additionally, we plan to enhance our classification experiments, as well as the
filling in of the messages' arguments, exploiting syntactic processing and
incorporating WordNet.\footnote{\url{http://www.cogsci.princeton.edu/~wn/}}

\section{Concluding Remarks}
This work has discussed the summarization of evolving events in terms of their
evolution in time --- linear, non-linear --- and the source --- synchronous,
asynchronous. Of course, we are not the first to introduce the notion of time
in summarization. \possessivecite{Allan&al01} work on temporal summarization is
such a case. In their work they take the results from a \tdt system for an
event, and they put all the sentences one after the other in chronological
order, regardless of the document that it belonged to, creating a stream of
sentences. Then they apply two statistical measures, \emph{usefulness} and
\emph{novelty}, to each ordered sentence. The aim is to extract those sentences
which have a score over a certain threshold. This approach differs from ours in
various ways. Firstly, they do not distinguish between the sources, while we
try to incorporate in our system the different viewpoints that the various
sources might have, and present them to the user. Also, they are not concerned
with the evolution of the events; instead they try to detect novel information.
Finally, we have an abstractive system, while they have an extractive one.

In terms of the source dimension, as far as we know, this has not been
discussed elsewhere.

Another point that should be stressed concerns the use of the cross-document
relations. In the past there have been several attempts to incorporate
relations, in one form or another, for the creation of a summary.
\citeN{Radev00}, for example, proposed the \cstlong which incorporated a set of
24 domain-independent relations that exist between various textual units across
documents. In a later paper \citeN{Zhang&al02} reduce that set to 17 relations
and perform experiments with human judges. Those experiments revealed several
interesting results. For example, human judges annotated only sentences,
ignoring the other textual units (phrases, paragraphs, documents) that the
theory suggests. Additionally, there was a rather small inter-judge agreement
concerning the type of relation that connects two sentences. Nevertheless,
\citeN{Zhang&al03} and \citeN{Zhang&Radev.04a} continue this work using \ml
algorithms to identify the cross-document relations. We have to note here that
although some cross-document relations such as agreement and disagreement might
be independent of the domain, we believe that in general cross-document
relations do depend on the domain. Another difference with our work is that our
relations concentrate on identifying the similarities and differences between
the sources, in two different axes: \emph{synchronically} and
\emph{diachronically}. In other words, we try to capture through those
relations the points of difference between the sources, as well as the
evolution of an event.

We are currently studying the summarization of non-linear events and extend our
summarization system in order to improve the performance of the extraction
sub-system.

\bibliography{nlucs05}
\bibliographystyle{achicago}

\end{document}